%% file: main.tex
\documentclass[11pt,twocolumn]{article}

\usepackage[margin=1in]{geometry}
\usepackage{amsmath,amssymb}
\usepackage[T1]{fontenc}
\usepackage{booktabs}
\usepackage{placeins}
\usepackage{graphicx}
\usepackage[hyphens]{url}
\usepackage{hyperref}
\usepackage{cite}
\usepackage{xcolor}
\usepackage{enumitem}
\usepackage{caption}
\usepackage{subcaption}

\hypersetup{colorlinks=true, linkcolor=blue, citecolor=blue, urlcolor=blue}

\newcommand{\sop}{spherical occupancy profile}
\newcommand{\sops}{spherical occupancy profiles}

\title{\bf Learning Spherical Occupancy Profiles for\\
  Multi-View 3D Reconstruction and Generation
  \thanks{Code and trained weights are available at
    \url{https://github.com/102324988/LSOP_code_release}}}

\author{\textbf{YiHsuan Tsai}\\
  Nanjing University\\
  \href{https://orcid.org/0009-0003-4532-077X}{\textsc{ORCID}: 0009-0003-4532-077X}}

\date{November 29, 2026}

\begin{document}
\raggedbottom
\maketitle

\begin{abstract}
Spherical occupancy profiles (SOPs) encode radial stopping weights
distilled from Gaussian fields fitted to multi-view images. We use
this representation in a ray-conditioned decoder and a generative
profile VAE with latent diffusion. On a fixed split of 999 Google
Scanned Objects, a matched comparison finds lower teacher-profile
Wasserstein error for a 96-bin SOP than for single-depth and
three-layer heads. The paired depth interval for SOP versus three
layers includes zero. The ray-conditioned decoder reaches a median
normalized teacher-relative depth error of $0.0351$ on 90 test
objects, versus $0.0384$ for its global-condition baseline. The
global baseline has lower Chamfer against independent test scans.
The generation experiments measure candidate spread, limited
held-out scan coverage, and outcomes of image-only selection.
The learned predictors use synthetic turntable views. A separate
two-scene DTU study measures the fixed geometric front end on real
photos.
\end{abstract}

\input{sections/intro.tex}
\input{sections/related.tex}
\input{sections/background.tex}
\input{sections/method.tex}
\input{sections/experiments.tex}

\input{sections/discussion.tex}
\input{sections/conclusion.tex}

\noindent\textbf{Declaration of generative AI and AI-assisted technologies in the writing process}\\[2pt]
During the preparation of this work the author used AI-assisted tools for drafting, language editing, and formatting. After using these tools, the author reviewed and edited the content as needed and takes full responsibility for the content of the publication.

\paragraph{Code and data availability.}
Checkpoints, object IDs, evaluation code, and per-object audit
results are available in the
\href{https://github.com/102324988/LSOP_code_release/tree/3d652bfe8a815421af89e4153294d2b71ac14323/results/audits}{versioned reproduction archive}.
The documentation lists the original renders and intermediate teacher
arrays required to rerun the data-dependent experiments.

\FloatBarrier
\bibliographystyle{plain}
\bibliography{refs}

\end{document}

%% file: sections/intro.tex
\section{Introduction}
\label{sec:intro}

Reconstructing 3D shape from multiple images is a long-standing problem in computer
vision, and the recent explosion of feed-forward reconstruction models~\cite{pragmatist,real3d,ermani2023,li2023lrm} and
diffusion-based generative priors~\cite{spgen,poole2022score,liu2023zero} has blurred the
line between \emph{discriminative} reconstruction (predict one shape from observations) and
\emph{generative} reconstruction (sample a shape consistent with observations). A central
design choice in both families is the \emph{intermediate representation}: occupancy grids,
signed distance fields, neural fields, and tri-planes are all used as the interface between
image evidence and the final surface.

We study the \emph{spherical occupancy profile} (\sop), a sequence of
transmittance-weighted stopping masses on each ray from the object center.
Fitting a Gaussian field to posed multi-view images and sampling its
Cartesian opacity proxy supplies reference profiles from images.
The regular angular grid and radial bins support both ray-conditioned
prediction and convolutional latent modeling. Profile location and shape
describe the teacher's radial surface evidence. Predicted peak values are
scores relative to a unit-peak training target.

We evaluate two branches on a fixed 819/90/90 split of 999 GSO objects
with 48 rendered views each. The discriminative branch combines
view-averaged features with image features sampled at projected ray
locations. At $\gamma=1.5$ it reaches $0.0351$ median normalized
soft-depth error on the held-out objects, versus $0.0384$ for the
global-condition decoder. We compare paired teacher-depth errors and evaluate the predicted clouds
against independent scans. A matched output-head experiment compares
one-depth, three-layer, and 96-bin radial representations. The generative branch uses a
profile VAE and latent diffusion to sample multiple candidate fields
conditional on images. We compare guidance settings by sample spread
and report reference-based best-of-eight selection as an oracle comparison.

Our morphology study compares learned and post-hoc sharpening at
specified values of $\gamma$. We report half-maximum width, peak score,
and depth readout on a common set of covered rays. We also audit the
fixed geometric front end on two calibrated DTU scenes against
structured-light references.

\begin{figure}[t]
  \centering
  \includegraphics[width=\linewidth]{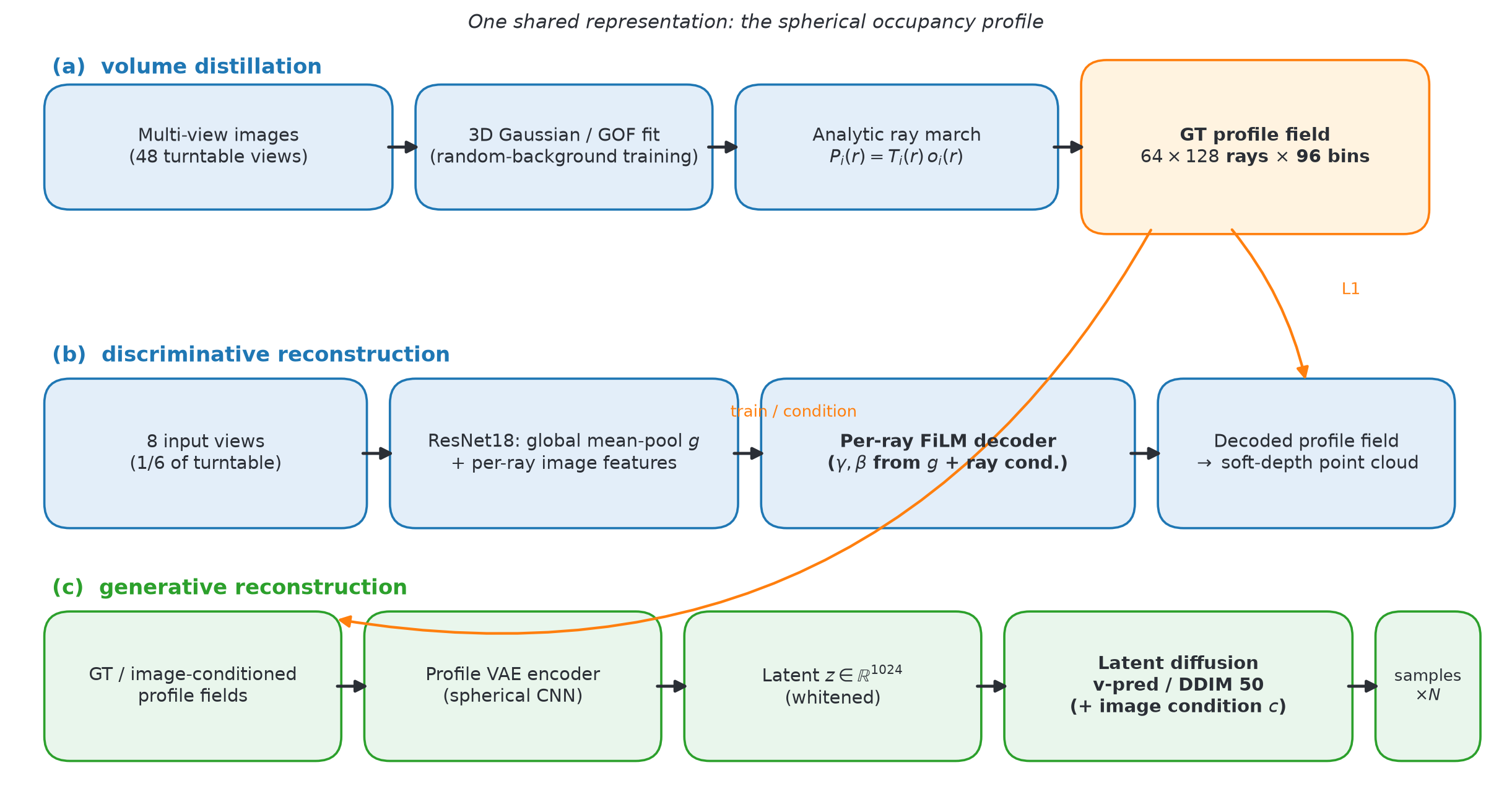}
  \caption{Pipeline overview. (a)~Volume distillation produces teacher-reference \sops{} from
  multi-view captures without mesh supervision; (b)~a discriminative per-ray decoder maps
  images to profile fields; (c)~a profile VAE with latent diffusion maps profile fields to
  a sampleable latent, enabling multi-solution reconstruction. The two branches share the
  same representation.}
  \label{fig:pipeline}
\end{figure}

%% file: sections/related.tex
\section{Related Work}
\label{sec:related}

\paragraph{Feed-forward 3D reconstruction.}
Feed-forward models predict a 3D representation from one or a few images in a single
forward pass. NeRF-based and tri-plane-based large reconstruction models
(LRM)~\cite{li2023lrm,ermani2023} popularized this paradigm, and subsequent works scaled it
to real photographs~\cite{real3d} or to sparse unposed multi-view
input~\cite{pragmatist}. A common limitation is that a feed-forward pass commits to a
single reconstruction, providing no mechanism for ambiguity or alternative hypotheses; recent
work partially addresses this by hallucination-aware diffusion priors that mask unreliable
predicted views during reconstruction~\cite{liu2026had}. Our discriminative branch follows the
same feed-forward philosophy but operates on per-ray occupancy profiles rather than view or
volume representations, and our generative branch is designed precisely to expose the
solution set that feed-forward models discard.

\paragraph{Neural fields, Gaussians, and radial representations.}
Neural radiance fields~\cite{mildenhall2020nerf} and 3D Gaussian
splatting~\cite{kerbl20233dgs} represent multi-view scenes. Gaussian
Opacity Fields~\cite{yu2024gof} fit a Gaussian field with ray-based
rendering and extract surfaces from it. We distill the fitted field
through a Cartesian opacity proxy and discrete radial stopping
weights. RayDF~\cite{liu2023raydf} learns ray--surface distances with
multi-view consistency. SPGen~\cite{spgen} represents shapes with
spherical layered depth and generates shapes from a single image.
RaySplats~\cite{byrski2025raysplats} uses analytic ray--Gaussian
intersections for rendering. Our experiment compares learned
depth layers and a shared soft radial profile under matched image
conditioning and teacher supervision.

\paragraph{Diffusion-based 3D generation.}
Score-distillation methods distill a 2D diffusion model into a 3D field~\cite{poole2022score},
and image-conditioned view diffusion~\cite{liu2023zero,shi2024mvdream} has become a standard
prior for reconstruction. Multi-view and panoramic diffusion~\cite{huang2024epidiff,
tang2024mvdiffhd,du2024diffpano} generate consistent image stacks, while depth-aware variants
operate on per-view depth maps~\cite{wang2024mvdd}. Closest to our generative branch are
methods that train a diffusion model in a compressed latent space~\cite{rombach2022ldm} on top
of a learned representation; SphereDiff~\cite{wang2025spherediff} does this for spherical
latents, and SPGen for spherical layered depth. Most closely related in architecture,
LN3Diff~\cite{lan2024ln3diff} learns a VAE over tri-plane neural fields and diffuses in that
latent for fast 3D generation, and SC-Diff~\cite{galvis2024scdiff} does the same in a discrete
TSDF latent for shape completion---both operate on plane or grid latents rather than on a
per-ray occupancy signal. We instead diffuse the latent space of a
\emph{profile VAE} and assess decoded coverage against original scans.

\paragraph{Multi-solution recon\-struction and uncer\-tainty.}
A growing body of work treats reconstruction as sampling from the posterior over 3D given
observations. Latent posterior sampling~\cite{chen2025latent} represents scenes as random
latents with a diffusion prior and draws multiple scene samples, using sample variance as an
uncertainty map; it operates in the tri-plane domain. Reconstruction-conditioned
diffusion~\cite{chang2025reconvviagen} couples an explicit reconstructor with a generative
prior, completing unseen parts while preserving the visible geometry. We measure the spread of candidate profiles at discrete settings
of classifier-free guidance~\cite{ho2022classifierfree}.
Complementary to posterior sampling, uncertainty has also been modeled explicitly for depth
estimation through Bayesian inference~\cite{kendall2017bayesian} and calibrated predictive
uncertainties in self-supervised monocular depth~\cite{poggi2020uncertainty}; our evaluation reports predicted peak scores and empirical coverage
of teacher depths.

%% file: sections/background.tex
\section{Background}
\label{sec:background}

\subsection{Gaussian fields and opacity}
A 3D Gaussian scene~\cite{kerbl20233dgs} represents a scene as a set of anisotropic
Gaussians, each with a mean $\mu$, a covariance $\Sigma$, an opacity $\alpha$, and a
view-dependent color. Rendering projects the Gaussians into a splatting rasterizer and
accumulates depth-ordered $\alpha$-blending, which makes the field differentiable with respect
to all parameters, so both geometry and appearance can be fitted from multi-view images by
gradient descent.

Gaussian Opacity Fields~\cite{yu2024gof} use ray-based volume
rendering and regularization to fit Gaussian fields and extract
surfaces. We use the fitted Gaussians as the teacher source.
Our distillation samples a Cartesian grid. At each grid point we
take the maximum of the filtered, opacity-weighted Gaussian
kernel values. This grid defines the opacity proxy used for radial
sampling.

\subsection{Spherical occupancy profiles}
For each canonical ray direction $\mathbf{d}_i$, we sample the fitted
GOF opacity grid at radii $r_j$ using trilinear interpolation. Let
$o_{ij}$ denote the sampled opacity proxy treated as attenuation per unit distance, $\Delta r$ the
radial spacing, and $a_{ij}=1-\exp(-o_{ij}\Delta r)$ the segment
opacity. The teacher derives a discrete stopping mass
\begin{equation}
  P_{ij}=T_{ij}a_{ij}, \qquad T_{ij}=\prod_{k<j}(1-a_{ik}),
  \label{eq:profile}
\end{equation}
where $T_{ij}$ is transmittance upon entering segment $j$.
Its sum is the teacher's ray coverage. Training targets normalize
each profile by its peak, so the learned profiles retain relative
radial shape while their absolute stopping mass is unspecified.

The angular samples form a regular $128\times64$ azimuth--polar grid.
Azimuth spans $[0,2\pi)$ with periodic wrap; the polar angle is sampled
over $[10^{-3},\pi-10^{-3}]$. The Cartesian grid contains $64^3$ samples over $[-1.3,1.3]^3$. The 96 radial samples span $[0,r_{\max}]$
with $r_{\max}\simeq2.252$ in the canonical object coordinates. The
radial spacing is about $0.024$. Reported ray-wise statistics use this
discrete grid, with each included ray contributing one sample.

\subsection{Data: GSO corpus}
We build our corpus on 999 objects of Google Scanned Objects (GSO), rendered on turntables
with 48 views each (3 elevation $\times$ 16 azimuth, $800\times600$, per-view random saturated
background). Each object is centered at its bounding-box center and
scaled to unit maximum vertex radius. The historical COLMAP export
used $f_x=(W/H)f_y$ although the rendered camera had square pixels.
For $800\times600$ images, it exported $1098.99$ and $824.24$ pixels.
The archived teachers and ray-conditioned checkpoints use that
export. Section~\ref{sec:discussion} reports the calibration audit.

For each of the 999 objects we (i)~train a GOF field from its 48 views (6{,}000 iterations,
2--3 GPU-minutes per object on an L20), and (ii)~sample the fitted field and compute the discrete \sop{} weights above. The result is a corpus of image stacks paired with profile fields.
We use a fixed split of the Google Scanned Objects (GSO) corpus~\cite{googlescannedobjects}: 819 training objects, 90 validation objects, and
90 held-out test objects (permutation seed 42). Later audits evaluate fixed archived checkpoints on these splits.

%% file: sections/method.tex
\section{Method}
\label{sec:method}

\subsection{Overview}
Our pipeline couples a \emph{discriminative} and a \emph{generative} branch on the same
\sop{} representation (Fig.~\ref{fig:pipeline}). The discriminative branch maps multi-view
images to profile fields directly; the generative branch maps images to the latent space of a
profile VAE and diffuses there, so that samples can be drawn and their spread quantified. Both
branches share the profile vocabulary of Sec.~\ref{sec:background}, and both are trained on
the volume-distilled corpus without mesh supervision.

\subsection{Discriminative reconstruction}
\label{sec:method:disc}
\textbf{Global condition.} A shared ResNet18 produces a
512-dimensional image feature for each of $K$ views. We concatenate
each feature with the sines and cosines of the view azimuth and
elevation, linearly project to 512 dimensions, and average over
views to obtain $g$.

\textbf{Per-ray decoder.} A 21-dimensional ray direction encoding
is projected to 64 dimensions and processed by three FiLM blocks
with affine modulation from $g$. A linear head and sigmoid produce
a 96-bin profile. For ray $i$ and bin $j$, let $p_{ij}$ be the prediction
and $P_{ij}$ the teacher stopping mass in Eq.~\eqref{eq:profile}.
Up to the batch average, the loss is
\begin{equation}
  \mathcal{L}_{\mathrm{prof}}
  = \frac{1}{|\mathcal{I}|B}
    \sum_{i\in\mathcal{I}}\sum_{j=1}^{B}w_i|p_{ij}-\hat{s}_{ij}|,
  \label{eq:v3loss}
\end{equation}
where $\mathcal{I}$ is the sampled ray set, $B=96$,
$w_i=\min(1,\sum_j P_{ij}+0.05)$, and $\hat{s}_{ij}$ divides $P_{ij}$
by its ray maximum with a numerical floor of $10^{-6}$.
For nonzero predictions, peak normalization is invariant to
positive rescaling. A loss on the normalized predictions is constant
along this scale direction. Equation~\eqref{eq:v3loss} supervises the
raw prediction on the unit-peak target scale. We call the resulting
global-condition model \textsc{d8}.

\textbf{Ray-specific image evidence.} Global conditioning discards per-ray evidence, which
we hypothesized to be the cause of systematically over-wide profiles. The refined model
(\textsc{m3b}) augments the decoder with a ray-specific pathway. For each ray direction
$\mathbf{d}_i$ on the sphere, we form a 3D point $\mathbf{X}_i = r_{\mathrm{ref}}\,\mathbf{d}_i$
at a fixed reference radius $r_{\mathrm{ref}}=1.0$, project it into each of the $K$ views with
the known camera poses, and bilinearly sample the ResNet18 layer-2/3/4 features at the projected pixel. The three scales
($128+256+512$ channels) are averaged over the $K$ views and concatenated into a
$896$-dimensional per-ray vector, which is projected to 128 dimensions and injected into the
FiLM stack as a \emph{second} conditioning channel $(\gamma_r,\beta_r)$, so the head is
modulated by both global and ray-specific image evidence. The model
inherits D8 weights and initializes the added modulation to zero. Training and inference use the fixed radius $r_{\mathrm{ref}}$.

\textbf{Learned sharpening.} We vary the target width by applying a power transformation to each ray:
\begin{equation}
  \hat{s}_{ij}^{(\gamma)} \;=\; \bigl(\hat{s}_{ij}\bigr)^{\gamma}, \qquad \gamma\in\{1.5,2\},
  \label{eq:gammatarget}
\end{equation}
applied per ray to the max-normalized teacher profile before the L1 loss.

\subsection{Generative reconstruction}
\label{sec:method:gen}
\textbf{Profile VAE.} We learn a VAE~\cite{kingma2013vae} over profile fields. The encoder applies a per-ray MLP
($96\to128\to64$) to each profile, arranges the resulting per-ray codes on the
$64\times128$ spherical grid, and processes them with three 2D convolutions
with circular azimuth padding and repeated polar boundary rows (channels $64\to128\to256\to256$), followed by global
pooling to a $1024$-dimensional latent $z$ with parameters $(\mu,\log\sigma^2)$. The decoder mirrors the
per-ray FiLM architecture of Sec.~\ref{sec:method:disc}, modulated by the global latent. The
objective is
\begin{equation}
  \begin{split}
    \mathcal{L}_{\mathrm{VAE}} \;=\; \mathbb{E}_{q(z|x)}
    \bigl[ \mathcal{L}_{\mathrm{prof}}\bigl(\mathrm{dec}(z), \hat{s}\bigr) \bigr] \\
    \qquad {} + \beta \, D_{\mathrm{KL}}\bigl(q(z|x)\,\|\,p(z)\bigr),
  \end{split}
  \label{eq:vaeloss}
\end{equation}
with $\beta{=}10^{-4}$ and a 5-epoch KL warm-up. The reconstruction loss is again the
profile L1 of Eq.~\eqref{eq:v3loss}.

\textbf{Latent diffusion.} We freeze the VAE and keep each
training object's posterior mean $\mu$. Each mean contributes one
unperturbed latent and eight independent perturbations
$\mu+0.5\epsilon$ with $\epsilon\sim\mathcal{N}(0,I)$. We whiten the
combined set per coordinate. Across 819 training objects, the
mean per-coordinate variance of the object means is $0.00370$,
or about $1.64\%$ of the archived augmented variance before
whitening. A time-conditioned MLP uses
$v$-prediction~\cite{ho2020ddpm,salimans2022vpred}.
Sampling uses 50 deterministic DDIM steps~\cite{song2021ddim}.

\textbf{Image-conditioned generation.} The frozen D8 pathway
aggregates image and angle features from the eight fixed views into
$c\in\mathbb{R}^{512}$. The velocity denoiser $v_\theta(z_t,c,t)$
learns with 10\% dropout of $c$ (\mbox{m3a\_cond}).
Classifier-free guidance~\cite{ho2022classifierfree} uses
$v_{\mathrm{u}}+w(v_{\mathrm{c}}-v_{\mathrm{u}})$, combining
predictions with zero condition and the observed condition.
The separate \mbox{m3a\_cond\_dd0} checkpoint uses zero dropout.

\textbf{Multi-solution sampling.} We draw $N$ conditioned samples for
each object. Pairwise Chamfer within their decoded clouds records
sample spread. The best-of-$N$ score selects the sample with the
lowest error against the teacher profile. We report this oracle result
alongside single-sample and averaged-profile readouts.

\subsection{Evaluation metrics and morphology}
\label{sec:method:metrics}
For each teacher-covered ray (coverage at least $0.02$), we compare
the predicted radial centroid or argmax bin with the teacher peak.
The per-object depth score is the median absolute error divided
by $r_{\max}$ and is denoted $\mathrm{dmed_s}$ or $\mathrm{dmed_h}$,
respectively. Split-level scores take the median over objects unless
specified otherwise.

For point-cloud evaluation we place one depth point on each retained
ray and compute the mean of the two directed nearest-neighbor mean
distances to the cloud lifted directly from the stored canonical teacher
peak radii. In the GSO discriminative
table this Chamfer is in the object's canonical coordinates, whose
bounding-sphere radius is one. Teacher coverage defines the support
in this evaluation. We report half-maximum width relative to the
teacher on covered rays whose teacher width spans at least two bins.
Predicted peak height is reported on the unit-peak target scale.

%% file: sections/experiments.tex
\section{Experiments}
\label{sec:experiments}

\subsection{Setup}
We use the GSO corpus of Sec.~\ref{sec:background} with the 819/90/90 split (seed 42).
All discriminative models take $K{=}8$ input views, randomly sampled per object during
training and fixed to frames $0,6,\dots,42$ at evaluation. The ray-specific
model uses the archived camera projection records described in
Section~\ref{sec:background}. The global-condition model \textsc{d8} trains for 40 epochs (Adam, lr $3\times10^{-4}$, batch 8,
StepLR $\times0.3$ at half epochs). The per-ray refinement \textsc{m3b} first freezes the
shared encoder and precomputes the global and per-ray features (a one-time pass over 909
objects), then trains only the ray-conditioned head (2.2M parameters) for 30 epochs (Adam,
lr $3\times10^{-4}$, batch 16, cosine to 5\%, grad-clip 1.0), initialized from \textsc{d8}
with the per-ray pathway inert. The profile VAE trains for 60 epochs at
$\beta{=}10^{-4}$ with a 5-epoch KL warm-up (Adam, lr $3\times10^{-4}$, StepLR). The latent
diffusion models train on the frozen VAE latent with $v$-prediction, a linear noise schedule
($\beta_1{=}10^{-4}\to\beta_T{=}2\times10^{-2}$, $T{=}1000$), 400 epochs (Adam, lr
$10^{-3}$, cosine to 3\%, grad-clip 1.0), and 50-step DDIM ($\eta{=}0$) sampling; the
image-conditioned variant additionally uses 10\% condition dropout during training. Metrics
are defined in Sec.~\ref{sec:method:metrics}. GSO depth errors use
$r_{\max}$ normalization. Discriminative GSO Chamfer distances use
canonical object coordinates after correcting the reference radii. The split, D8 checkpoint, and supplement
outputs are available in the code repository.

\subsection{Discriminative reconstruction}
\label{sec:exp:disc}
Table~\ref{tab:disc} reports validation and held-out test errors.
The global-condition D8 model has the lowest validation depth median
of the listed image-conditioned models. The $\gamma=1.5$ M3b variant
reduces the median normalized teacher-depth error from 0.0384 to
0.0351. We recompute the teacher-cloud Chamfer from archived
predictions and peak depths. Those depths already encode canonical
radii. The historical D5 pack multiplied the reference radii by
$r_{\max}$ a second time, affecting the original Chamfer values.
The corrected test Chamfer is 0.05826 for D8 and 0.05415 for M3b.
The median paired M3b-minus-D8 difference is $-0.00302$ with a
95\% object-bootstrap interval of $[-0.00482,-0.00182]$.

On 90 held-out objects, $\gamma=1.5$ improves depth error on 58
objects in a retrospective fixed-checkpoint comparison. Table~\ref{tab:disc}
reports the archived variants. The median of paired M3b-minus-D8 differences is $-0.00431$
with a 95\% object-bootstrap percentile interval of
$[-0.00543,-0.00061]$ (30,000 paired resamples).
The difference between the two marginal medians is $-0.00333$
with interval $[-0.00788,0.00111]$. These intervals describe
variation across objects at fixed checkpoints.

\begin{table}[t]
  \centering
  \footnotesize
  \setlength{\tabcolsep}{3pt}
  \caption{Discriminative reconstruction (median over objects; lower is better).
  $\mathrm{dmed}$ = soft-depth error, $\mathrm{ch}$ = Chamfer to the teacher-peak cloud in canonical coordinates.}
  \label{tab:disc}
  \begin{tabular}{lcccc}
    \toprule
    method & \multicolumn{2}{c}{$\mathrm{dmed_s}$} & \multicolumn{2}{c}{$\mathrm{ch_s}$} \\
     & val & test & val & test \\
    \midrule
    \textsc{d8} (global cond.) & 0.0338 & 0.0384 & 0.05773 & 0.05826 \\
    \textsc{m3b} v1 (per-ray L1) & 0.0364 & 0.0356 & 0.05830 & 0.05492 \\
    \textsc{m3b} v2 $\gamma{=}1.5$ & 0.0354 & 0.0351 & 0.05712 & 0.05415 \\
    \textsc{m3b} v2 $\gamma{=}2$ & 0.0348 & 0.0356 & 0.05808 & 0.05311 \\
    \textsc{m3b} v2, rdim 512 & 0.0354 & 0.0347 & 0.05793 & 0.05371 \\
    \bottomrule
  \end{tabular}
\end{table}

\begin{figure}[t]
  \centering
  \includegraphics[width=\linewidth]{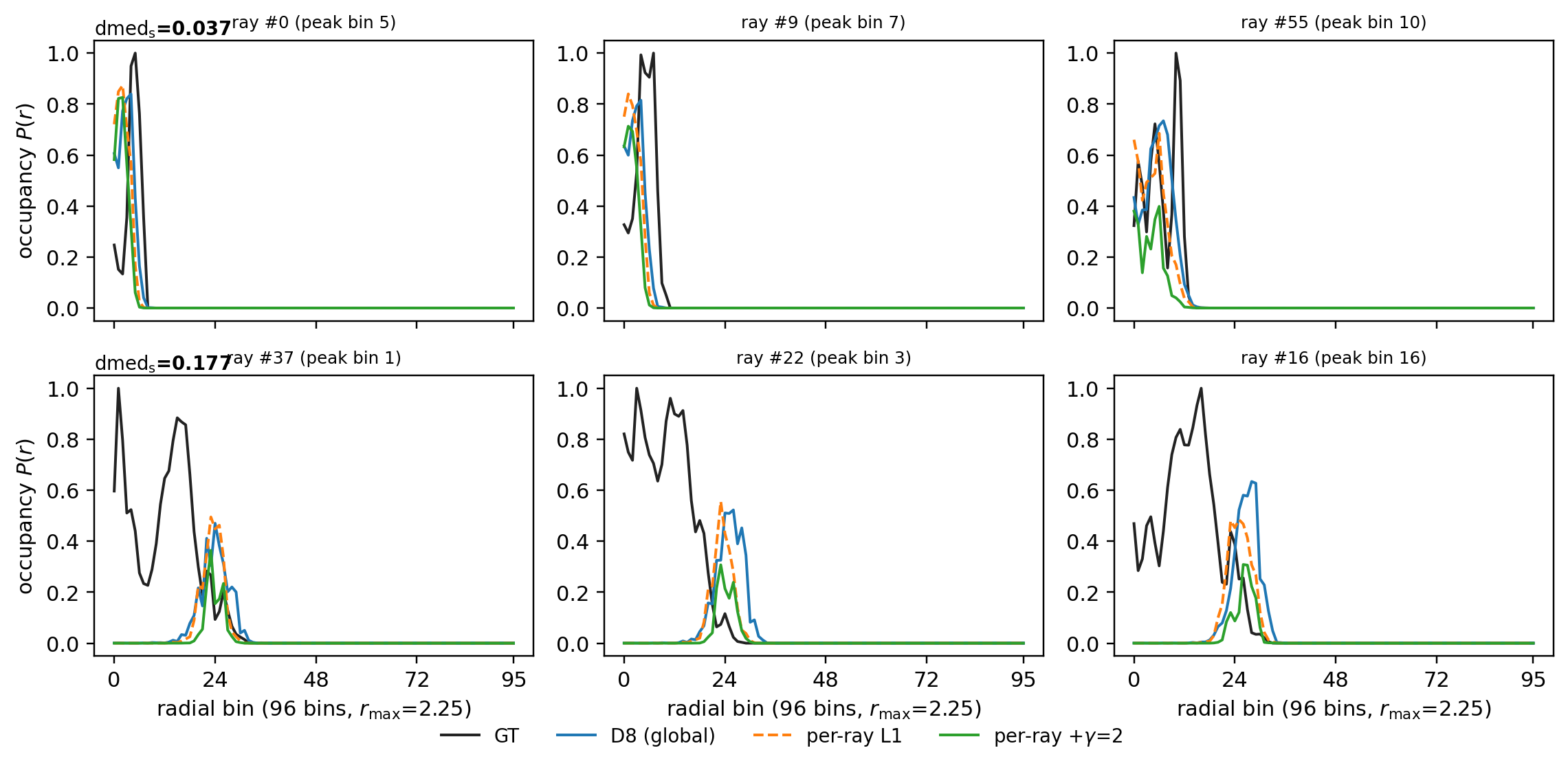}
  \caption{Example ray profiles for a mid- and a hard object. Teacher profiles (black) are narrow
  with peak $\approx1$; \textsc{d8} (blue) is over-wide and low; the per-ray model (orange) is
  sharper; the sharpened-target model (green, $\gamma{=}2$) approaches the teacher width in these examples.
  Three covered rays (peak bins at the 25/50/75 percentiles) are shown per object.}
  \label{fig:profiles}
\end{figure}

Figure~\ref{fig:profiles} illustrates two archived examples. The
teacher profiles peak at approximately one after normalization.
In the illustrated rays, the per-ray variant and the sharpened
target yield narrower responses than the global decoder.

\paragraph{Center robustness.}
The spherical parameterization is anchored at the object center; in practice the center comes
from a detector or background subtraction, so we measure tolerance to mis-centering. We shift all
8 input views horizontally by $\Delta x$ pixels (reflect-padded, keeping content in frame) and
re-evaluate \textsc{d8} on the 90-object validation split. Objects span $\approx 96$\,px in the
224\,px input, so shifts of $\{2,4,9\}$\,px equal $\{4.2,8.3,18.7\}$\% of the projected object
radius. Table~\ref{tab:center} shows soft-depth error stays within 11\% of baseline up to
$\approx 19\%$ radius mis-centering ($0.0338\to0.0373$): this measures sensitivity to the applied image shift. (Ground-truth profiles are referenced to the nominal center, so this
measures input-misalignment tolerance, not center re-estimation.)

\begin{table}[t]
  \centering
  \footnotesize
  \setlength{\tabcolsep}{4pt}
  \caption{Center-robustness of \textsc{d8} (90-object validation split). Shifting all 8 views by
  $\Delta x$ px simulates an error in the object-center estimate of the spherical
  parameterization.}
  \label{tab:center}
  \begin{tabular}{lcccc}
    \toprule
    $\Delta x$ [px] & shift / obj.\ radius & $\mathrm{dmed_s}$ & rel.\ to baseline \\
    \midrule
    0 & 0\% & 0.0338 & 1.00 \\
    2 & 4.2\% & 0.0338 & 1.00 \\
    4 & 8.3\% & 0.0351 & 1.04 \\
    9 & 18.7\% & 0.0373 & 1.10 \\
    \bottomrule
  \end{tabular}
\end{table}

\subsection{Inference efficiency}
\label{sec:exp:eff}
Table~\ref{tab:eff} lists fixed-input timings on an NVIDIA L20 in fp32.
The D8 model takes 2.2\,ms for its timed GPU stages and 1.6\,ms per
object at batch eight. The M3b GPU stages total 4.85\,ms. Its CPU
ray-grid construction was logged separately at 2.38\,ms, giving a
serial sum of about 7.23\,ms before input and output overheads.
M1 uses a teacher profile as input and takes 1.5\,ms for its timed
stages. One M2 unconditional 50-step DDIM sample and decode take
19.5\,ms. These are stage timings. The teacher's per-object GOF fit
is a separate processing stage.

\begin{table*}[t]
  \centering
  \footnotesize
  \setlength{\tabcolsep}{4pt}
  \caption{Median single-object inference latency and peak GPU memory (NVIDIA L20, fp32). forward
  = view encoding (\textsc{d8}/\textsc{m3b}), latent encoding (M1), or 50-step DDIM (M2); dec =
  per-ray decoding of the $8192$-ray profile field.}
  \label{tab:eff}
  \begin{tabular}{lcccc}
    \toprule
    method & forward [ms] & dec [ms] & total [ms] & peak mem [MB] \\
    \midrule
    \textsc{d8} (1 object) & 1.5 & 0.7 & 2.2 & 165 \\
    \textsc{d8} (batch 8) & --- & --- & 1.6/obj & 525 \\
    \textsc{m3b} v2 (1 object) & 3.6 & 1.2 & 4.8\textsuperscript{a} & 474 \\
    VAE recon (M1) & 0.6 & 0.9 & 1.5 & 241 \\
    latent diff.\ sample (M2) & 18.5 & 1.0 & 19.5 & 239 \\
    \bottomrule
  \end{tabular}
  {\footnotesize \textsuperscript{a} GPU stages only. Separate CPU ray-grid setup took 2.38\,ms.}
\end{table*}

\subsection{External baseline and view scaling}
\label{sec:exp:ext}
We compare D8 with a single-distance head inspired by RayDF~\cite{liu2023raydf}.
The models share a ResNet18 backbone family, mean-pool fusion, data split
and evaluation support. They differ in output, supervision and coverage
weighting. The distance head regresses one teacher depth per ray, while
D8 predicts the radial profile. At 8 and 1 views the distance head
has higher validation medians, but lower test medians (0.0369 versus
0.0384 at 8 views and 0.0422 versus 0.0442 at one view). These are
descriptive comparisons of fixed runs with the view protocol in
Table~\ref{tab:ext}.

\begin{table}[t]
  \centering
  \footnotesize
  \setlength{\tabcolsep}{4pt}
  \caption{External baseline and view scaling on the same split and protocol. RW = ray-wise
  distance field re-implemented in the spirit of RayDF~\cite{liu2023raydf}, sharing the
  encoder, fusion, optimizer, and metric with \textsc{d8}. $\mathrm{dmed_s}$ = median absolute
  soft-depth error over covered rays (objects); lower is better.}
  \label{tab:ext}
  \begin{tabular}{lcccc}
    \toprule
    model & views $K$ & val & test \\
    \midrule
    \textsc{d8} (profile) & 8 & 0.0338 & 0.0384 \\
    \textsc{d8} (profile) & 2 & 0.0430 & 0.0419 \\
    \textsc{d8} (profile) & 1 & 0.0380 & 0.0442 \\
    RW (RayDF-style) & 8 & 0.0396 & 0.0369 \\
    RW (RayDF-style) & 1 & 0.0455 & 0.0422 \\
    \bottomrule
  \end{tabular}
\end{table}

The frozen D8 checkpoint accepts varying numbers of input views
because its fusion uses mean pooling. On the evaluated fixed view
sets, the test errors at $K=1,2,8,16$ are 0.0412, 0.0445, 0.0384
and 0.0374, respectively. The results depend on the number and
placement of the chosen views.

We also jitter the azimuth and elevation encodings with Gaussian noise
while holding the image and ray grid fixed. Medians range from
0.0384 to 0.0385 for the five-seed sweep through $30^\circ$.
With one seed and 30 test objects, the $0^\circ$, $90^\circ$ and
$180^\circ$ standard-deviation settings give 0.0444, 0.0423 and
0.0434. On a 60-object subset, rotating the image in its plane
through $8^\circ$ gives 0.0378--0.0386. These interventions probe
D8 sensitivity to input metadata and appearance at the measured
settings.

\begin{table}[t]
  \centering
  \footnotesize
  \setlength{\tabcolsep}{4pt}
  \caption{Deployment sensitivity of the frozen $K{=}8$ \textsc{d8} model (mean-pool fusion),
  test split, $\mathrm{dmed_s}$ (lower is better). Left: view-count scaling at inference.
  Center: Gaussian pose-encoding noise of standard deviation $\sigma$ (five seeds). Right:
  image-plane roll of each input view.}
  \label{tab:q4}
  \begin{tabular}{cc@{\qquad}cc@{\qquad}cc}
    \toprule
    views $K$ & $\mathrm{dmed_s}$ & $\sigma$ & $\mathrm{dmed_s}$ & roll & $\mathrm{dmed_s}$ \\
    \midrule
    1  & 0.0412 & $6^\circ$  & 0.0384 & $0^\circ$ & 0.0384 \\
    2  & 0.0445 & $10^\circ$ & 0.0384 & $2^\circ$ & 0.0383 \\
    8  & 0.0384 & $15^\circ$ & 0.0385 & $4^\circ$ & 0.0382 \\
    16 & 0.0374 & $20^\circ$ & 0.0385 & $6^\circ$ & 0.0386 \\
       &        & $30^\circ$ & 0.0384 & $8^\circ$ & 0.0378 \\
    \bottomrule
  \end{tabular}
\end{table}

\subsection{Profile morphology and the width--peak frontier}
\label{sec:morph}
\begin{figure}[t]
  \centering
  \includegraphics[width=\linewidth]{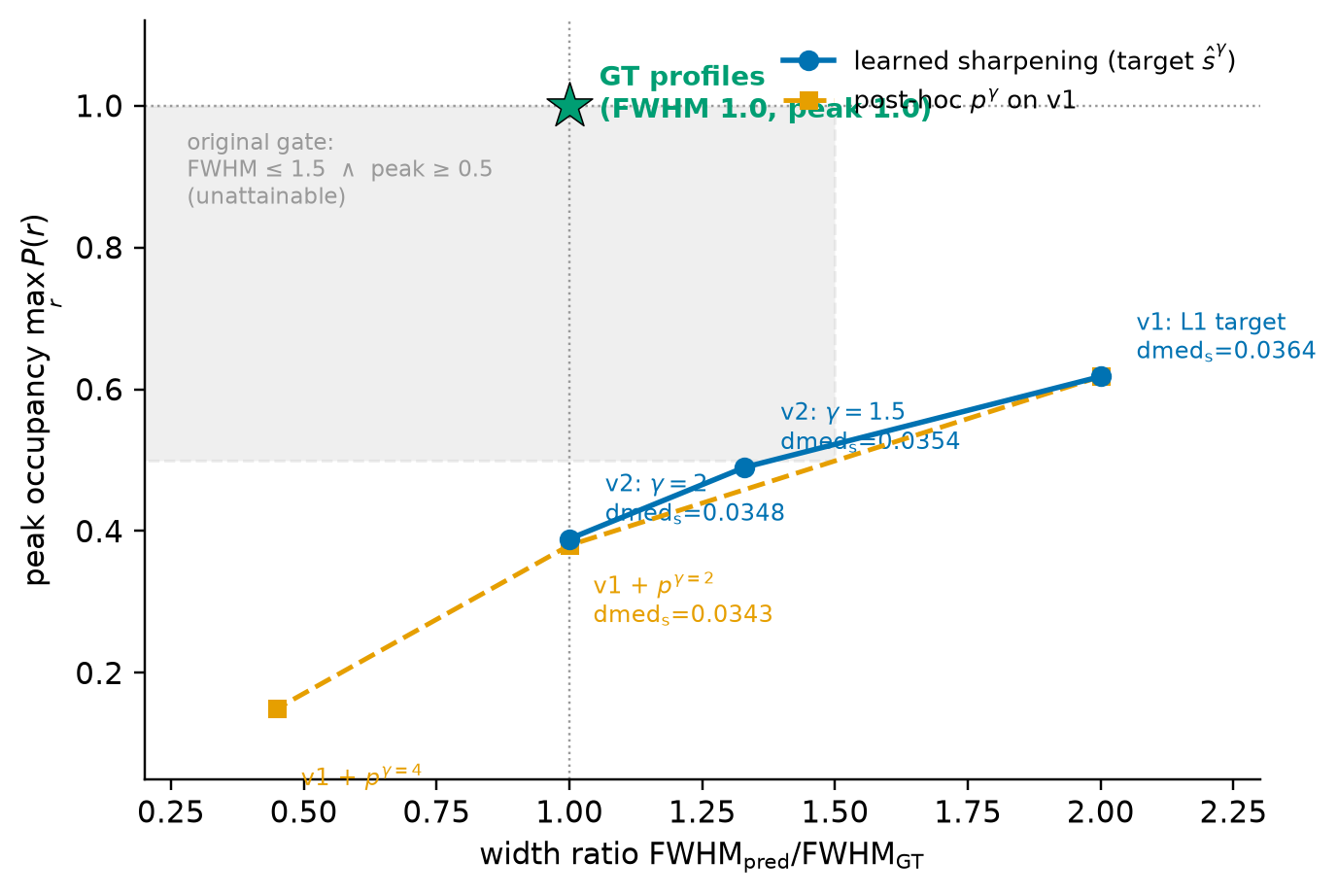}
  \caption{Morphology frontier on the validation split. Both learned sharpening
  ($\hat{s}^\gamma$, blue) and post-hoc $p^\gamma$ (orange) reduce width and peak score at the measured settings. The grey box shows the reference cutoffs.}
  \label{fig:morph}
\end{figure}

Table~\ref{tab:morph} and Fig.~\ref{fig:morph} summarize the morphology analysis. We report the
profile width ratio to the teacher and the peak score, with soft-depth error as a
reference.

At the measured settings, sharpening reduces median width and
peak score. Training with $\gamma=2$ produces a median FWHM ratio
of 1.00; on test objects its interquartile range is [0.50, 1.67].
Post-hoc $p^2$ produces a validation ratio of 1.00. In Table~\ref{tab:morph}
$\gamma=2$ improves soft depth on validation relative to unsharpened
M3b v1, and the two have similar test depth medians. We treat the
width--peak pattern as an empirical property of the measured
settings. Width and amplitude are reported separately because their
interpretations depend on the profile scale.

\begin{table}[t]
  \centering
  \small
  \caption{Profile morphology (val). FWHM = predicted-to-teacher width ratio (teacher $=1.0$);
  peak is the maximum predicted score relative to a unit-peak target.}
  \label{tab:morph}
  \begin{tabular}{lccc}
    \toprule
    method & FWHM & peak & $\mathrm{dmed_s}$ \\
    \midrule
    Teacher profiles & 1.00 & 1.00 & --- \\
    \textsc{d8} (global cond.) & 2.00 & 0.59 & 0.0338 \\
    \textsc{m3b} v1 (per-ray L1) & 2.00 & 0.62 & 0.0364 \\
    \textsc{m3b} v2 $\gamma{=}1.5$ & 1.33 & 0.49 & 0.0354 \\
    \textsc{m3b} v2 $\gamma{=}2$ & 1.00 & 0.39 & 0.0348 \\
    v1 + post-hoc $p^{\gamma=2}$ & 1.00 & 0.38 & 0.0343 \\
    v1 + post-hoc $p^{\gamma=4}$ & 0.45 & 0.15 & --- \\
    \bottomrule
  \end{tabular}
\end{table}

\subsection{Generative reconstruction and multi-solution behavior}
\label{sec:exp:gen}
For M1, the archived validation depth median is 0.0419.
An archived unconditional M2 run with 16 samples has a median
generated pairwise distance of 0.0941 versus 0.1498 across
reference objects. The corresponding median widths in bins are
nine for generated samples, seven for VAE reconstructions, and
four for reference profiles. These summaries describe the sampled
shapes and their spread.

The single-sample M3a validation run gives 0.0489. The
90-object spread run in Table~\ref{tab:spread} groups the samples
per object. Its spread at $w=1$ is 18\% of the inter-object reference,
and the tested guidance settings give different spreads.

\begin{table*}[t]
  \centering
  \small
  \caption{Archived generative depth readouts on validation objects.}
  \label{tab:gen}
  \begin{tabular}{lc}
    \toprule
    model or readout & $\mathrm{dmed_s}$ \\
    \midrule
    VAE reconstruction (M1) & 0.0419 \\
    conditional diffusion, single sample (M3a) & 0.0489 \\
    \bottomrule
  \end{tabular}
\end{table*}

\begin{table*}[t]
  \centering
  \small
  \caption{Multi-solution spread vs.\ classifier-free guidance (M3a, val). Spread =
  intra-object pairwise Chamfer of $N{=}8$ samples; inter-object reference 0.1194.}
  \label{tab:spread}
  \begin{tabular}{lcc}
    \toprule
    CFG weight $w$ & spread & rel.\ to inter-object \\
    \midrule
    0 (unconditioned) & 0.0820 & 69\% \\
    1 & 0.0217 \ $[0.014,0.040]$ & 18\% \\
    8 (strong) & 0.0445 & 37\% \\
    \bottomrule
  \end{tabular}
\end{table*}

\subsection{Real photographs: DTU}
\label{sec:exp:dtu}
We audit the fixed geometric front end on two calibrated DTU
scenes, scan1 and scan6, using 49 photos per scene and structured-light
reference points. Table~\ref{tab:dtu} separates directed errors and
their bidirectional mean. The reconstruction-to-scan medians are
3.98 and 4.62\,mm. The respective 90th percentiles are 35.85 and
139.42\,mm, reflecting a substantial tail for scan6.

\begin{table*}[t]
  \centering
  \small
  \caption{Real-photo front-end audit on DTU (GOF 20{,}000 iters).}
  \label{tab:dtu}
  \begin{tabular}{lcc}
    \toprule
    metric & scan1 & scan6 \\
    \midrule
    silhouette overlap & 1.000 & 1.000 \\
    cross-view depth, $\mathrm{p50}$ & 0.0063 & 0.011 \\
    consistency (10\%) & 0.90 & 0.75 \\
    floaters & 0 & 0 \\
    \midrule
    pred. $\to$ scan, median [mm] & 3.98 & 4.62 \\
    pred. $\to$ scan, p90 [mm] & 35.85 & 139.42 \\
    scan $\to$ pred., mean [mm] & 1.27 & 1.64 \\
    scan coverage at 5\,mm [\%] & 99.96 & 98.6 \\
    bidirectional mean Chamfer [mm] & 7.92 & 23.31 \\
    \bottomrule
  \end{tabular}
\end{table*}

\input{sections/revision_audits.tex}

%% file: sections/revision_audits.tex
\subsection{Independent scan audit}
We compare predictions with the original GSO meshes for all 90 test
objects. The archived bounding-box center and scale match the retrieved
source meshes. We sample 20,000 surface points per scan and use the
canonical coordinate frame without post-hoc alignment. The primary
comparison uses teacher coverage. We also threshold each model's own
peak at 0.30 as a sensitivity check. Empty predictions are counted in
the reported denominators and excluded from finite point-cloud
medians.

\begin{table*}[t]
  \centering
  \small
  \caption{Independent scan audit on 90 held-out GSO objects. All
    distances are in the canonical frame. F1 is computed per object
    at distance 0.02 and then summarized by its median.}
  \label{tab:scan_audit}
  \begin{tabular}{llrrr}
    \toprule
    model & support & finite clouds & median Chamfer & median F1 \\
    \midrule
    D8 & teacher coverage & 90 & 0.16601 & 0.04627 \\
    M3b $\gamma=1.5$ & teacher coverage & 90 & 0.16967 & 0.03086 \\
    D8 & predicted peak $\geq0.30$ & 85 & 0.16651 & 0.03441 \\
    M3b $\gamma=1.5$ & predicted peak $\geq0.30$ & 87 & 0.17687 & 0.02475 \\
    \bottomrule
  \end{tabular}
\end{table*}

The median paired M3b-minus-D8 Chamfer difference is $0.00681$
with a 95\% object-bootstrap interval of $[0.00236,0.01218]$ under
teacher coverage. Under predicted support it is $0.00836$
with interval $[0.00482,0.01391]$ on the 84 pairs with two nonempty
clouds. The teacher-relative depth benefit therefore coexists with
higher Chamfer in this independent comparison.

\subsection{Generated distribution audit}
For each of three seeds we sample 90 latent codes from M2 using
50-step DDIM and 90 from a standard Gaussian in the same whitened
latent coordinates. The models share inverse whitening, the decoder,
and the peak threshold 0.15. We compare their clouds with 90
independent test scans using minimum matching distance (MMD) and
coverage~\cite{achlioptas2018pointclouds}. Here MMD averages each
reference's closest finite generated-cloud Chamfer. Coverage is
the fraction of reference shapes matched as the nearest neighbor
of at least one generated cloud.

Across seeds, median MMD is 0.2306 for M2 and 0.2470 for the
matched Gaussian. Each model covers 2 of 90 references at each seed.
M2 yields 67--73 finite clouds per run, compared with 90 for the
Gaussian. At this readout the measured reduction in matching
distance accompanies low reference coverage. In the other direction,
the median generated-to-nearest-test distance across seeds is 0.13868
for M2 and 0.13270 for the Gaussian baseline. A separate CPU draw
compares nearest distances to the same 90 test scans and 90
training scans selected by object-ID hash from the verified cached
meshes. Both models are closer to the test pool for every finite
sample in this check. All nearest test matches concentrate on two
references. This pool-level diagnostic leaves memorization of
individual training shapes unresolved.

To test the augmentation scale, we refit two unconditional denoisers
per seed for 400 epochs using the same archived means, frozen decoder,
and optimizer schedule. The augmentation standard deviations are
$0.5$ and $0.05$. At each of three paired seeds both settings still
cover 2 of 90 test scans with 90 draws. Across seeds, the median MMD
is 0.22348 with $0.5$ and 0.23661 with $0.05$. Reducing the
perturbation alone leaves coverage low under this fixed decoder and
point-cloud readout.

\subsection{Image-only candidate ranking}
Eight samples are grouped by their conditioning object for each
guidance setting. We evaluate an image-based selector that extracts
foreground from the input colors, projects each candidate cloud into
the eight known views, and ranks candidates by median silhouette
intersection over union. Camera geometry is tied to the perspective
renderer. No target profile or scan is consulted during selection.

Both validation and test use all 90 objects in their original
split and the same eight original rendered views per object.
Before the remaining originals were recovered, an input sensitivity
run used 10 original validation and 16 original test render sets
with the other views recreated from the renderer recipe and pinned
source meshes. On two overlapping objects, the minimum PSNR over
16 paired views was 51.6\,dB. That sensitivity run led to the same
qualitative conclusions.

\begin{table*}[t]
  \centering
  \small
  \caption{Image-only selection and reference-based oracle with eight
    candidates per object. Coverage is the fraction of covered teacher
    depths between the empirical 5th and 95th percentiles of eight
    candidate depths, summarized by the median object.}
  \label{tab:image_rank}
  \begin{tabular}{lrrrrr}
    \toprule
    split & guidance & first sample & image selected & oracle & interval coverage \\
    \midrule

    val & 1 & 0.0521 & 0.0541 & 0.0415 & 7.7\% \\
    val & 8 & 0.0555 & 0.0626 & 0.0377 & 24.1\% \\
    test & 1 & 0.0490 & 0.0538 & 0.0411 & 11.5\% \\
    test & 8 & 0.0520 & 0.0623 & 0.0386 & 24.2\% \\
    \bottomrule
  \end{tabular}
\end{table*}

On test at $w=8$, the paired median difference between image
selection and the first sample is $0.00193$, with a 95\% object-bootstrap
interval of $0.00000$ to $0.00865$. The tested selector shows no improvement,
and its sample interval covers 24.2\% of teacher depths at the median
test object. With eight continuous draws and one target, jointly
exchangeable, even the min--max interval covers the target with
probability $7/9$. Here the empirical percentiles describe candidate
spread and agreement with a fixed teacher under this protocol.

We extend the same per-object draws to 64 on the original test images
at $w=8$. Coverage is 33.8\%, with a paired median change of
3.7 percentage points and a 95\% object-bootstrap interval
of $[1.6, 5.4]$ points. With 64 draws, the median test object still has
43.4\% of its covered teacher depths outside the full sample range.

\subsection{Comparison with discrete radial layers}
To isolate compression error from image conditioning, we quantize
the teacher profiles for all 90 test objects. A single depth is
the radial median. Three equal-mass layers use the 1/6, 1/2 and
5/6 quantiles. We compare these empirical distributions with the
original teacher using the integrated absolute CDF difference
on the normalized radial axis. The median object mean drops
from 0.03922 to 0.01457. The paired median change is $-0.02450$
with 95\% interval $[-0.02630,-0.02171]$. Against the teacher
peak, the corresponding depth errors from the compressed
distributions are 0.01053 and 0.02105. This teacher-informed
comparison measures radial representation error from direct encoding.

We separately train three heads on the original 819/90/90 split using
eight original views per object. The heads share the same frozen D8
image encoder, three-layer ray backbone, AdamW optimizer, and sum of
normalized centroid-depth L1 and integrated absolute radial CDF
error. The output parameterizations are one learned depth, three
learned depths with learned mass fractions, and a 96-bin SOP. The
depth heads splat their masses onto neighboring bins. Each
head trains for 12 epochs at seeds 7, 19 and 42. We retain up to
3,072 valid rays per training object and draw 768 per visit. The
evaluation uses all teacher-covered rays with nonzero target mass.

\begin{table*}[t]
  \centering
  \small
  \caption{Matched head comparison. Entries are medians across three
    seeds of the object medians. Each object's depth error is the
    median absolute error over covered rays and its radial $W_1$ is
    the mean over those rays. Depth and radius use the normalized
    radial coordinate.}
  \label{tab:matched_heads}
  \begin{tabular}{lrrrr}
    \toprule
    representation & val depth & val radial $W_1$ & test depth & test radial $W_1$ \\
    \midrule
    Single depth & 0.03637 & 0.04981 & 0.03940 & 0.05260 \\
    Three layers & 0.03613 & 0.03812 & 0.03977 & 0.04039 \\
    SOP (96 bins) & 0.03632 & 0.03523 & 0.03946 & 0.03428 \\
    \bottomrule
  \end{tabular}
\end{table*}

On test, SOP yields the lowest radial $W_1$ in each seed. Depth
medians across seeds span 0.03582--0.04104 for one depth, 0.03822--0.04302 for three layers,
and 0.03570--0.03959 for SOP. Averaging each object's paired differences over
seeds and taking the median across objects, SOP minus three layers
gives $-0.00377$ ($[-0.00521,-0.00227]$) in radial $W_1$ and $-0.00044$ ($[-0.00227,+0.00078]$) in depth (95\% object-bootstrap
intervals). These intervals quantify object sampling variation for
the fitted heads. The output dimensions differ, and radial $W_1$
is a training term. The audit measures fit to the teacher profiles
with a frozen synthetic-view conditioner.

In a repeat training with the same settings, we stratify covered rays by
teacher profile shape. The three-layer head varies between GPU runs;
both trainings give lower radial error for SOP at each seed. The separated-peak
group requires two interior maxima of at least 20\% of the ray maximum,
at least six bins apart, with a valley at or below half the smaller peak.
An object contributes to a group when at least 32 of its rays qualify.
The group contains 90 test objects, and 90 objects contribute to
the remaining-ray group. For SOP minus three layers, the paired radial
$W_1$ change on separated-peak rays is $-0.00481$ ($[-0.00576,-0.00388]$) and the depth change is $-0.00056$ ($[-0.00130,0.00135]$).
The paired radial change on the remaining-ray group is $-0.00325$ ($[-0.00490,-0.00247]$). Intervals are
95\% object-bootstrap intervals. This stratification uses teacher peaks as
an operational definition of profile shape.

%% file: sections/discussion.tex
\section{Discussion}
\label{sec:discussion}

\paragraph{Geometry, depth, and scope.}
Relative to the global-condition D8 model, M3b improves the typical
paired test-object depth error. The canonical teacher-cloud comparison
also favors M3b $\gamma=1.5$ on test. Validation depth summaries vary
across the variants. The single-distance head has
slightly lower test depth medians at both matched view counts in
Table~\ref{tab:ext}. The independent scan comparison in
Table~\ref{tab:scan_audit} gives D8 lower median Chamfer than M3b under
both support rules. This points to distinct effects on teacher-relative
depth and the predicted surface against original scans. The matched head
comparison in Table~\ref{tab:matched_heads} finds lower radial
distribution error for the SOP output. The paired depth interval
for SOP versus three layers includes zero. The teacher-peak strata characterize this result by profile shape
and leave physical occlusion as a separate question.

\paragraph{Synthetic camera calibration.}
The historical GSO intrinsic export set $f_x=(W/H)f_y$ although
the perspective renderer used equal horizontal and vertical focal
lengths in pixel coordinates. The export passed to the COLMAP
initialization and ray-to-pixel projection. In a check on 16
original-render test objects and eight views each, the median
object silhouette IoU is 0.994 for the renderer intrinsics and
0.741 for the export. The paired median IoU change is 0.253
with a 95\% percentile interval of [0.246, 0.289]. At inference, we also change only the horizontal focal length of
the ray-to-pixel projection for the released, frozen M3b checkpoint.
On the 90 original-image test objects, the paired depth change is
$0.00004$ with a 95\% interval $[-0.00009,0.00028]$. Independent
scan Chamfer on teacher support increases by a paired median of
$0.00238$ ($[0.00075,0.00420]$). In this rerun, the historical arm
has a median per-object profile MAE of $0.01388$ against the archived
predictions and a depth median of $0.0370$. The paired sensitivity
uses the two arms of the rerun. A clean estimate of the training
effect requires refitting GOF and retraining with corrected cameras.

\paragraph{Profile shape and score scale.}
The measured sharpening settings narrow the median profiles. An
elementwise power transformation on $[0,1]$ with exponent greater
than one decreases fractional peaks. Positive rescaling of a ray
preserves its half-maximum width and ideal centroid, so a joint
width--peak threshold also depends on the chosen amplitude scale.
We report shape and peak separately. Calibrating either peak scores
or sample intervals to geometric errors requires an external target.

\paragraph{Conditional generation.}
Guidance affects sample spread in the tested runs. On original inputs,
the image-only silhouette selector yields higher test depth medians than
the first sample at both guidance settings (Table~\ref{tab:image_rank}).
At $w=8$, the median interval coverage is 33.8\% with 64 samples.
The independent unconditional audit finds low reference coverage for
both the diffusion model and its matched Gaussian baseline.

\paragraph{Real photos and known cameras.}
The DTU study audits the fixed front end on two calibrated scenes.
The learned predictors use object-centered turntable images with
known cameras. The image-shift, roll, and metadata perturbations
measure distinct sensitivities of D8. Translation and rotation error in the ray-to-pixel projection remain open. The centered
grid is intended for isolated objects.

%% file: sections/conclusion.tex
\section{Conclusion}
\label{sec:conclusion}
A radial stopping-weight representation distilled from Gaussian fields
supports an image-conditioned profile predictor and a generative latent
model. The matched output-head experiment finds lower radial
distribution error for SOP, and the paired depth interval includes
zero. The ray-conditioned refinement improves teacher-relative depth,
while D8 has lower Chamfer against independent GSO scans. The
generation audits find low reference coverage, no observed gain from
the tested image selector, and limited agreement between candidate
intervals and teacher depths. The two-scene DTU audit characterizes
the fixed geometric front end.

%% file: refs.bib
@inproceedings{mildenhall2020nerf,
  title     = {{NeRF}: Representing Scenes as Neural Radiance Fields for View Synthesis},
  author    = {Mildenhall, Ben and Srinivasan, Pratul P. and Tancik, Matthew and Barron, Jonathan T. and Ramamoorthi, Ravi and Ng, Ren},
  booktitle = {European Conference on Computer Vision (ECCV)},
  year      = {2020}
}

@inproceedings{kerbl20233dgs,
  title     = {3D Gaussian Splatting for Real-Time Radiance Field Rendering},
  author    = {Kerbl, Bernhard and Kopanas, Georgios and Leimk{\"u}hler, Thomas and Drettakis, George},
  booktitle = {ACM SIGGRAPH Conference Proceedings},
  year      = {2023}
}

@inproceedings{yu2024gof,
  title     = {Gaussian Opacity Fields: Efficient and Compact Surface Reconstruction in Unbounded Scenes},
  author    = {Yu, Zehao and Chen, Anpei and Huang, Binbin and Sattler, Torsten and Geiger, Andreas},
  booktitle = {ACM SIGGRAPH Asia Conference Proceedings},
  year      = {2024}
}

@inproceedings{liu2023raydf,
  title     = {RayDF: Neural Ray-surface Distance Fields with Multi-view Consistency},
  author    = {Liu, Zhuoman and Zhang, Yangxintong and Zhou, Zizhao and Lin, Jiabin and Yang, Bo},
  booktitle = {IEEE/CVF International Conference on Computer Vision (ICCV)},
  year      = {2023}
}

@article{li2023lrm,
  title   = {{LRM}: Large Reconstruction Model for Single Image to 3D},
  author  = {Li, Jiahao and Tan, Hao and Xu, Kai and Li, Zexiang and Liang, Yucheng and Chen, Yongyi and Zheng, Shangzhe and Cai, Yujun and Fan, Xiaolong and Li, Yu and Xia, Yuda and Wang, Hao and Liu, Jing and Li, Zhengyang and Li, Ye and Zhang, Ke and Yuan, Zhenjie and Xu, Yujun and Xu, Yuehao and Zhang, Xiaoyu and Fu, Lele and Yang, Ziwei and Liu, Yujun and Zhang, Jingbo and Fu, Cong and Cui, Zhe and Zeng, Siyuan and Zhang, Yuxuan and Sun, Zheng},
  journal = {arXiv preprint arXiv:2311.04400},
  year    = {2023}
}

@inproceedings{ermani2023,
  title     = {Triplane Meets Gaussian Splatting: Fast and Generalizable Single-View 3D Reconstruction with Transformers},
  author    = {Zou, Zi-Xin and Yu, Zhipeng and Guo, Yuan-Chen and Li, Yangguang and Liang, Ding and Cao, Yan-Pei and Zhang, Song-Hai},
  booktitle = {IEEE/CVF Conference on Computer Vision and Pattern Recognition (CVPR)},
  year      = {2024}
}

@inproceedings{real3d,
  title     = {{Real3D}: Scaling up Large Reconstruction Models with Real-World Images},
  author    = {Jiang, Haoyu and others},
  booktitle = {IEEE/CVF International Conference on Computer Vision (ICCV)},
  year      = {2025}
}

@inproceedings{pragmatist,
  title     = {Pragmatist: Multiview Conditional Diffusion Models for High-Fidelity 3D Reconstruction from Unposed Sparse Views},
  author    = {Wang, Songwen and Zhao, Qixuan and Hu, Kaiyue and Lan, Yilin and Liu, Linqi},
  booktitle = {AAAI Conference on Artificial Intelligence (AAAI)},
  year      = {2025}
}

@article{liu2026had,
  title   = {Hallucination-Aware Diffusion for Robust Multi-View 3D Reconstruction},
  author  = {Liu, Xiaofeng and others},
  journal = {arXiv preprint arXiv:2605.16873},
  year    = {2026}
}

@inproceedings{spgen,
  title     = {Spherical Layered Depth Synthesis with Diffusion Priors},
  author    = {Zeng, Yanan and others},
  booktitle = {ACM SIGGRAPH Asia},
  note      = {arXiv:2509.12721},
  year      = {2025}
}

@inproceedings{poole2022score,
  title     = {DreamFusion: Text-to-3D using 2D Diffusion},
  author    = {Poole, Ben and Jain, Ajay and Barron, Jonathan T. and Mildenhall, Ben},
  booktitle = {International Conference on Learning Representations (ICLR)},
  year      = {2023}
}

@inproceedings{liu2023zero,
  title     = {Zero-1-to-3: Zero-shot One Image to 3D Object},
  author    = {Liu, Ruoshi and Wu, Rundi and Van Hoorick, Basile and Tokmakov, Pavel and Zakharov, Sergey and Vondrick, Carl},
  booktitle = {IEEE/CVF International Conference on Computer Vision (ICCV)},
  year      = {2023}
}

@inproceedings{shi2024mvdream,
  title     = {{MVDream}: Multi-view Diffusion for 3D Generation},
  author    = {Shi, Yichun and Wang, Peng and Ye, Jianglong and Mai, Long and Li, Kejie and Liu, Xiao},
  booktitle = {International Conference on Learning Representations (ICLR)},
  year      = {2024}
}

@inproceedings{huang2024epidiff,
  title     = {{EpiDiff}: Enhancing Multi-View Synthesis via Localized Epipolar-Constrained Diffusion},
  author    = {Huang, Zehuan and Wen, Hao and Dong, Junting and Wang, Songcen and Yang, Yang and Duan, Ling-Yu and Guo, Guodong},
  booktitle = {IEEE/CVF Conference on Computer Vision and Pattern Recognition (CVPR)},
  year      = {2024}
}

@inproceedings{tang2024mvdiffhd,
  title     = {{MVDiffHD}: A Dense High-resolution Multi-view Diffusion Model for Single or Sparse-view 3D Object Reconstruction},
  author    = {Tang, Shitao and Chen, Jiacheng and Wang, Dilin and Tang, Chengzhou and Zhang, Fuyang and Fan, Yuying and Chandra, Vikas and Furukawa, Yasutaka and Ranjan, Rakesh},
  booktitle = {European Conference on Computer Vision (ECCV)},
  year      = {2024}
}

@inproceedings{du2024diffpano,
  title     = {{DiffPano}: Scalable and Consistent Text-to-Panorama Generation with Spherical Epipolar-Aware Diffusion},
  author    = {Du, Weicai and Chen, Ye and Wan, Hongyu and Peng, Yifan and Zhang, Tao and Li, Chenfeng and Chen, Wenyi and Zhang, Hongxiang and Liu, Hao and Guo, Jinzhu and Xu, Wei and Liu, Wei and Luo, Peng and Yu, Weihua and Shi, Jing and He, Ke and Xu, Yuanchun and Zeng, Jie and Wang, Zun and Wang, Wenyu and Zhou, Zeyu and Zhang, Jing and Ye, Mao and Zhu, Siyu and Wu, Feng},
  booktitle = {IEEE/CVF Conference on Computer Vision and Pattern Recognition (CVPR)},
  year      = {2024}
}

@inproceedings{wang2024mvdd,
  title     = {{MVDD}: Multi-View Diffusion Distillation for High-Fidelity Novel View Synthesis},
  author    = {Wang, Jinsong and Gao, Rui and Zhang, Wenqi and others},
  booktitle = {European Conference on Computer Vision (ECCV)},
  note      = {arXiv:2312.04875},
  year      = {2024}
}

@inproceedings{rombach2022ldm,
  title     = {High-Resolution Image Synthesis with Latent Diffusion Models},
  author    = {Rombach, Robin and Blattmann, Andreas and Lorenz, Dominik and Esser, Patrick and Ommer, Bj{\"o}rn},
  booktitle = {IEEE/CVF Conference on Computer Vision and Pattern Recognition (CVPR)},
  year      = {2022}
}

@article{wang2025spherediff,
  title   = {{SphereDiff}: Spherical Equivariant Latent Diffusion for 3D Generation},
  author  = {Wang, Tianyu and others},
  journal = {arXiv preprint arXiv:2504.14396},
  note    = {AAAI 2026 Oral},
  year    = {2025}
}

@inproceedings{chen2025latent,
  title     = {Predicting 3D Structure by Latent Posterior Sampling},
  author    = {Chen, Bohan and others},
  booktitle = {International Conference on Learning Representations (ICLR)},
  year      = {2025}
}

@inproceedings{ho2022classifierfree,
  title     = {Classifier-Free Diffusion Guidance},
  author    = {Ho, Jonathan and Salimans, Tim},
  booktitle = {NeurIPS Workshop on Deep Generative Models and Downstream Applications},
  year      = {2021}
}

@inproceedings{salimans2022vpred,
  title     = {Progressive Distillation for Fast Sampling of Diffusion Models},
  author    = {Salimans, Tim and Ho, Jonathan},
  booktitle = {International Conference on Learning Representations (ICLR)},
  year      = {2022}
}

@inproceedings{ho2020ddpm,
  title     = {Denoising Diffusion Probabilistic Models},
  author    = {Ho, Jonathan and Jain, Ajay and Abbeel, Pieter},
  booktitle = {Advances in Neural Information Processing Systems (NeurIPS)},
  year      = {2020}
}

@inproceedings{song2021ddim,
  title     = {Denoising Diffusion Implicit Models},
  author    = {Song, Jiaming and Meng, Chenlin and Ermon, Stefano},
  booktitle = {International Conference on Learning Representations (ICLR)},
  year      = {2021}
}

@inproceedings{kingma2013vae,
  title     = {Auto-Encoding Variational Bayes},
  author    = {Kingma, Diederik P. and Welling, Max},
  booktitle = {International Conference on Learning Representations (ICLR)},
  year      = {2014}
}

@inproceedings{kendall2017bayesian,
  title     = {What Uncertainties Do We Need in Bayesian Deep Learning for Computer Vision?},
  author    = {Kendall, Alex and Gal, Yarin},
  booktitle = {Advances in Neural Information Processing Systems (NeurIPS)},
  year      = {2017}
}

@inproceedings{poggi2020uncertainty,
  title     = {On the Uncertainty of Self-Supervised Monocular Depth Estimation},
  author    = {Poggi, Matteo and Aleotti, Filippo and Tosi, Fabio and Mattoccia, Stefano},
  booktitle = {IEEE/CVF Conference on Computer Vision and Pattern Recognition (CVPR)},
  year      = {2020}
}

@misc{googlescannedobjects,
  title        = {Google Scanned Objects: A High-Quality Dataset of 3D Objects},
  author       = {{Google Research / Everyday Robots}},
  howpublished = {Google Scanned Objects dataset, available at \url{https://research.google/pubs/google-scanned-objects-a-high-quality-dataset-of-3d-scanned-household-items/}},
  year         = {2022},
  note         = {Accessed 2026}
}

@inproceedings{lan2024ln3diff,
  title     = {{LN3Diff}: Scalable Latent Neural Fields Diffusion for Speedy 3D Generation},
  author    = {Lan, Yushi and Hong, Fangzhou and Yang, Shuai and Zhou, Shangchen and Meng, Xuyi and Dai, Bo and Pan, Xingang and Loy, Chen Change},
  booktitle = {European Conference on Computer Vision (ECCV)},
  note      = {arXiv:2403.12019},
  year      = {2024}
}

@article{galvis2024scdiff,
  title   = {{SC-Diff}: 3D Shape Completion with Latent Diffusion Models},
  author  = {Galvis, Juan D. and Zuo, Xingxing and Schaefer, Simon and Leutenegger, Stefan},
  journal = {arXiv preprint arXiv:2403.12470},
  year    = {2024}
}

@article{byrski2025raysplats,
  title   = {RaySplats: Ray Tracing based Gaussian Splatting},
  author  = {Byrski, Krzysztof and Mazur, Marcin and Tabor, Jacek and Dziarmaga, Tadeusz and K{\k{a}}dzio{\l}ka, Marcin and Baran, Dawid and Spurek, Przemys{\l}aw},
  journal = {arXiv preprint arXiv:2501.19196},
  year    = {2025}
}

@article{chang2025reconvviagen,
  title   = {ReconViaGen: Towards Accurate Multi-view 3D Object Reconstruction via Generation},
  author  = {Chang, Jiahao and Luo, Zhongjin and Ye, Chongjie and Wu, Yushuang and Chen, Yuantao and Zhang, Yidan and Li, Chenghong and Zhi, Yihao and Han, Xiaoguang},
  journal = {arXiv preprint arXiv:2510.23306},
  year    = {2025}
}

@inproceedings{achlioptas2018pointclouds,
  title = {Learning Representations and Generative Models for 3D Point Clouds},
  author = {Achlioptas, Panos and Diamanti, Olga and Mitliagkas, Ioannis and Guibas, Leonidas J.},
  booktitle = {Proceedings of the 35th International Conference on Machine Learning},
  pages = {40--49},
  year = {2018}
}
